\documentclass[letterpaper]{article} 
\usepackage{aaai2027}  
\usepackage[hyphens]{url}  
\usepackage{graphicx} 
\usepackage{natbib}  
\usepackage{caption} 
\usepackage{algorithm}
\usepackage{algorithmic}
\usepackage{xspace}
\newcommand{\method}{CORE\xspace}
\usepackage{textcomp}
\usepackage{booktabs}
\usepackage{graphicx}
\usepackage{multirow}
\usepackage{amsmath,amssymb}
\usepackage{float}
\usepackage{booktabs}
\usepackage{multirow}
\usepackage[table]{xcolor}
\definecolor{AvgRow}{HTML}{F2F2FF}
\usepackage{float}      
\usepackage{placeins}   

\usepackage{pdfpages}

\usepackage{cuted}

\usepackage{newfloat}
\usepackage{listings}
\DeclareCaptionStyle{ruled}{labelfont=normalfont,labelsep=colon,strut=off} 
\floatstyle{ruled}
\newfloat{listing}{tb}{lst}{}
\floatname{listing}{Listing}

\usepackage{booktabs}

\title{CORE: Contrast-and-Reflect for Self-Supervised Prompt Optimization in Anomaly Synthesis}
\author{
    Haoxiang Rao\textsuperscript{\rm 1},
    Fang Zhao\textsuperscript{\rm 1$\dagger$},
    Zhao Wang\textsuperscript{\rm 2},
    Weijia Li\textsuperscript{\rm 1},
    Chenyang Si\textsuperscript{\rm 1},
    Caifeng Shan\textsuperscript{\rm 1$\dagger$}
}
\affiliations{
    \textsuperscript{\rm 1}School of Intelligence Science and Technology, Nanjing University
    \quad
    \textsuperscript{\rm 2}China Mobile Zijin Innovation Institute
}

\usepackage{etoolbox}
\makeatletter
\newcommand{\COREtightentitle}{%
    \patchcmd{\@maketitle}
        {\vskip 0.1in plus 0.5fil minus 0.05in}
        {\vskip 18pt}{}{}%
    \patchcmd{\@maketitle}
        {\vskip 1em plus 2fil}
        {\vskip 20pt}{}{}%
}
\COREtightentitle
\COREtightentitle
\makeatother

\begin{document}

\maketitle

\renewcommand{\thefootnote}{\fnsymbol{footnote}}
\footnotetext[2]{Corresponding Authors.}
\renewcommand{\thefootnote}{\arabic{footnote}}

\begin{abstract}
Industrial anomaly detection (AD) is bottlenecked by the scarcity of anomalous samples, making high-fidelity anomaly synthesis crucial for improving downstream performance. However, existing methods either rely on scarce real anomaly references or auxiliary data, while lacking complex reasoning and iterative optimization capabilities for generating realistic defects in open-set scenarios.
To address these challenges, we propose \textbf{\method}, a cost-efficient and training-free prompt optimization framework based on Large Multimodal Models (LMMs), which unifies zero-shot and one-shot anomaly synthesis. \method first introduces an Anomaly Context Grounding module to transform sparse user context into defect-aware generation context, which is subsequently instantiated as structured anomaly prompts that disentangle normal-region preservation from anomaly-specific variation, enabling controllable and faithful defect generation. 
Furthermore, \method autonomously refines prompts through a self-supervised contrast-and-reflect mechanism. This design empowers general-purpose LMMs to synthesize realistic anomalies under both zero-shot and one-shot settings without task-specific training.
Extensive experiments on MVTec-AD and VisA demonstrate that \method synthesizes realistic and diverse anomalies, yielding superior improvements in downstream AD performance. Furthermore, its model-agnostic and plug-and-play paradigm enables effortless integration with diverse LMMs, facilitating adaptation to unseen open-set scenarios.\end{abstract}

\section{Introduction}

\begin{figure*}[t]
    \centering
    \includegraphics[width=\textwidth]{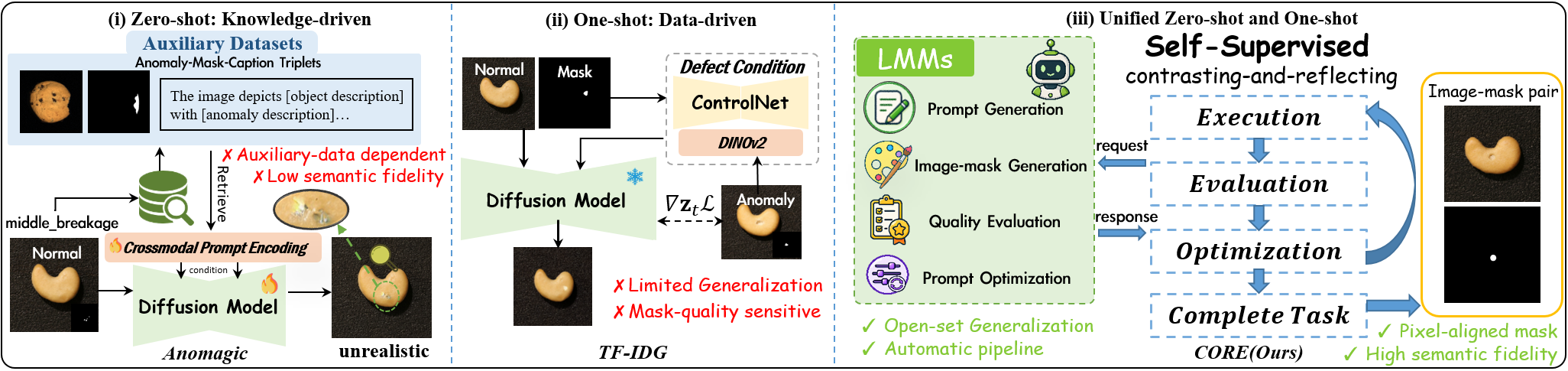}
    \caption{
    Comparison of representative anomaly synthesis paradigms, where existing (i) zero-shot and (ii) one-shot methods are limited by their reliance on auxiliary datasets and scarce anomaly references, respectively, while (iii) CORE unifies both settings through self-supervised closed-loop optimization for realistic and generalizable anomaly synthesis.
    }
    \label{fig:P1}
\end{figure*}

Industrial anomaly detection (AD) plays a crucial role in inspection and quality control by identifying, localizing, and categorizing visual defects~\cite{liu2024deep,huang2025deep}. However, industrial data suffer from severe data imbalance, where normal samples are abundant while anomalous samples are scarce. Although models trained~\cite{li2024promptad,guo2025dinomaly} solely on normal data can detect distributional deviations, their ability to localize and classify anomalies in complex scenarios remains limited due to insufficient modeling of real defect patterns. This motivates anomaly synthesis as an effective means to enrich defect distributions and improve supervised AD. Therefore, synthesizing realistic and diverse anomalies has emerged as a key strategy to enhance supervised AD.

Anomaly synthesis methods can be broadly categorized into zero-shot and few-shot settings according to the availability of real anomalous samples. Zero-shot methods are attractive for defect-scarce industrial scenarios~\cite{zhang2024realnet,sun2025unseen}, but early cut-paste strategies~\cite{li2021cutpaste,zavrtanik2021draem} often lack structural consistency and contextual realism. Recent diffusion-based approaches improve visual quality by leveraging pretrained generative priors, e.g., anomaly-related attention strengthen~\cite{sun2025unseen} or text-guided style transfer~\cite{so2026anostyler}. However, such internal guidance provides limited anomaly-specific knowledge, while Anomagic~\cite{jiang2026anomagic} retrieves relevant anomalous pairs from purpose-built auxiliary data, as shown in Fig.~\ref{fig:P1}(i), making its performance highly dependent on data quality and retrieval accuracy. These dependencies introduce substantial computational and data-preparation overhead, hindering efficient adaptation to open-set scenarios.

Few-shot methods leverage real anomalies to improve synthesis fidelity, yet training-based approaches are prone to overfitting when anomalies are extremely scarce~\cite{hu2024anomalydiffusion,dai2025seas}. More practical one-shot training-free methods~\cite{xu2025training,rao2026one}, as illustrated in Fig.~\ref{fig:P1}(ii), use a single anomaly reference to guide generation, yet rely on anomaly masks for defect placement and remain sensitive to spatially implausible locations. Moreover, their limited ability to model logical and abstract anomalies restricts coverage of complex defects. These limitations motivate a unified framework capable of synthesizing realistic and diverse anomalies across zero-shot and one-shot settings.

Recent advances in Large Multimodal Models (LMMs) have demonstrated strong open-world visual understanding, reasoning, and generation capabilities~\cite{wu2025omnigen2,deng2025emerging,seedream2025seedream}. Although LMMs have been explored for anomaly detection~\cite{li2023myriad,gu2024anomalygpt}, their potential for industrial anomaly synthesis remains underexplored. Due to limited domain-specific knowledge and weak sensitivity to fine-grained local defects, directly using general-purpose LMMs often fails to stably generate realistic, localized, and semantically accurate anomalies. To address these limitations, we propose \textbf{CORE}, a cost-efficient and training-free LMM-based framework that achieves high-fidelity anomaly synthesis via self-supervised prompt optimization.

As illustrated in Fig.~\ref{fig:P1}(iii), CORE provides a unified training-free framework for both zero-shot and one-shot anomaly synthesis. In industrial inspection, anomalies typically manifest as local deviations from otherwise normal products. Accordingly, we formulate anomaly synthesis as an image editing task to introduce plausible defect patterns into a given normal image without modifying its non-anomalous content. To enable controllable anomaly editing, CORE first grounds available visual and semantic contexts into defect-aware generation context, which is then instantiated as structured prompts. Following the normal-anomaly separability principle~\cite{duan2025anomalycontrol,rao2026one}, CORE decomposes the generation prompt into three complementary parts: BASE to preserve normal content, VARIANT to specify anomaly-related variations, and NEGATIVE to suppress undesired artifacts and irrelevant modifications. To further improve generation quality, CORE introduces a contrasting-and-reflecting mechanism inspired by LLM-as-a-judge studies~\cite{zheng2023judging,wang2024promptagent,yuksekgonul2025optimizing}, forming a closed-loop optimization process.

In the zero-shot setting, the key challenge is optimizing anomaly prompts without access to real anomalous references. CORE addresses this challenge by observing that prompt quality can be reflected through the generated results: a well-designed anomaly prompt should simultaneously achieve semantic consistency, local defect realism, and faithful preservation of non-anomalous regions. Instead of relying on human annotations or real anomaly references, CORE derives self-supervised optimization signals by contrasting candidates generated from different prompts. The comparison results expose prompt-level weaknesses in semantic consistency, local realism, and preservation of normal regions, and the resulting reflective feedback is used to revise weaker prompts. In the one-shot setting, CORE incorporates a single real anomaly reference as a weak visual anchor for category-level realism rather than a template for direct replication. This enables the model to capture defect appearance and material interaction while avoiding overfitting to the reference’s specific shape, location, or extent. In this way, CORE treats both generated outputs and limited anomaly references as feedback sources for prompt refinement, enabling realistic and diverse anomaly synthesis without task-specific training.

Our main contributions are summarized as follows:
\begin{itemize}
    \item We propose CORE, a cost-efficient and training-free unified framework for both zero-shot and one-shot industrial anomaly synthesis, reducing reliance on real anomalies, finetuning, and auxiliary datasets.

    \item We introduce a contrasting-and-reflecting mechanism that converts generated results and limited references into self-supervised feedback for prompt optimization.
    
    \item Experiments on MVTec-AD and VisA demonstrate that CORE achieves superior generation quality and downstream AD performance, while enabling plug-and-play transfer across LMM backends and strong generalization to unseen objects in open-world scenarios.
\end{itemize}

\section{Related Work}

\subsection{Anomaly Generation}
\noindent\textbf{Few-shot Anomaly Generation.}
Few-shot anomaly generation synthesizes diverse anomalies from limited real samples in the target category. Recent methods have evolved from GAN-based generation~\cite{goodfellow2020generative,duan2023few} to diffusion models with stronger generative capacity. Some approaches freeze the diffusion backbone and learn anomaly-specific embeddings~\cite{hu2024anomalydiffusion,gui2024few,lu2025generate,gal2023an}, while others~\cite{jin2025dual,song2025defectfill,dai2025seas,tong2026cadiff} finetune the backbone following DreamBooth personalization~\cite{ruiz2023dreambooth}. However, when anomalous samples are extremely scarce, training-based methods are prone to overfitting and unreliable distribution modeling, motivating training-free anomaly generation. Among training-free methods, TF-IDG~\cite{xu2025training} guides anomaly appearance by minimizing the distribution discrepancy between generated and real defects, while O2MAG~\cite{rao2026one} transfers anomaly patterns by manipulating multiple diffusion processes via self-attention grafting. Nevertheless, both methods are sensitive to spatial plausibility of anomaly masks and provide limited support for logical anomalies.

\noindent\textbf{Zero-shot Anomaly Generation.}
Zero-shot anomaly generation aims to synthesize realistic defects for unseen categories without requiring anomalous samples. Early cut-paste methods~\cite{li2021cutpaste,zavrtanik2021draem} overlay external textures onto normal images, but lack structural consistency and contextual realism. AnomalyAny~\cite{sun2025unseen} enhances anomaly-related token responses by modifying the cross-attention mechanism~\cite{vaswani2017attention} of Stable Diffusion, yet token-level guidance remains limited in capturing complex anomaly semantics. Anomagic~\cite{jiang2026anomagic} introduces cross-modal prompts and auxiliary retrieval, but its performance depends on auxiliary data quality and retrieval accuracy. AnoStyler~\cite{so2026anostyler} formulates anomaly generation as text-guided style transfer and fine-tunes a stylization network at test time. Overall, these methods remain constrained by limited anomaly-specific guidance and struggle to generate realistic defects.

\subsection{Large Multimodal Models (LMMs)}
Large Multimodal Models (LMMs)~\cite{hurst2024gpt,team2023gemini} have shown strong capabilities in open-world visual understanding, reasoning and generation. Representative systems such as Gemini~\cite{team2023gemini} and GPT~\cite{hurst2024gpt} demonstrate the potential of general-purpose multimodal models as flexible generators conditioned on visual and textual inputs. While recent studies have explored LMMs for industrial anomaly detection~\cite{zhang2025towards,li2026iad}, their ability to synthesize realistic industrial anomalies remains underexplored. Since LMMs are trained on broad Internet-scale data and lack domain-specific defect priors, directly using them may cause inaccurate anomaly semantics or unintended changes to non-anomalous regions. CORE addresses this by treating LMMs as black-box backends and performing self-supervised prompt optimization through a contrast-and-reflect mechanism.

\section{Method}
We formulate anomaly synthesis as a constrained image editing task, with the goal of autonomously optimizing the anomaly prompt \(P\) to generate anomalous image-mask pairs. To bridge the domain knowledge gap between general-purpose Large Multimodal Models (LMMs) and industrial anomaly synthesis while minimizing human intervention, we propose \textbf{CORE}, a self-supervised framework that constructs and optimizes anomaly prompts through a contrast-and-reflect mechanism. CORE derives setting-specific feedback signals to guide prompt optimization under both zero-shot and one-shot synthesis scenarios. The following subsections introduce the overall framework, structured prompt formulation, and prompt optimization mechanism.

\begin{figure*}[t]
    \centering
    \includegraphics[width=\textwidth]{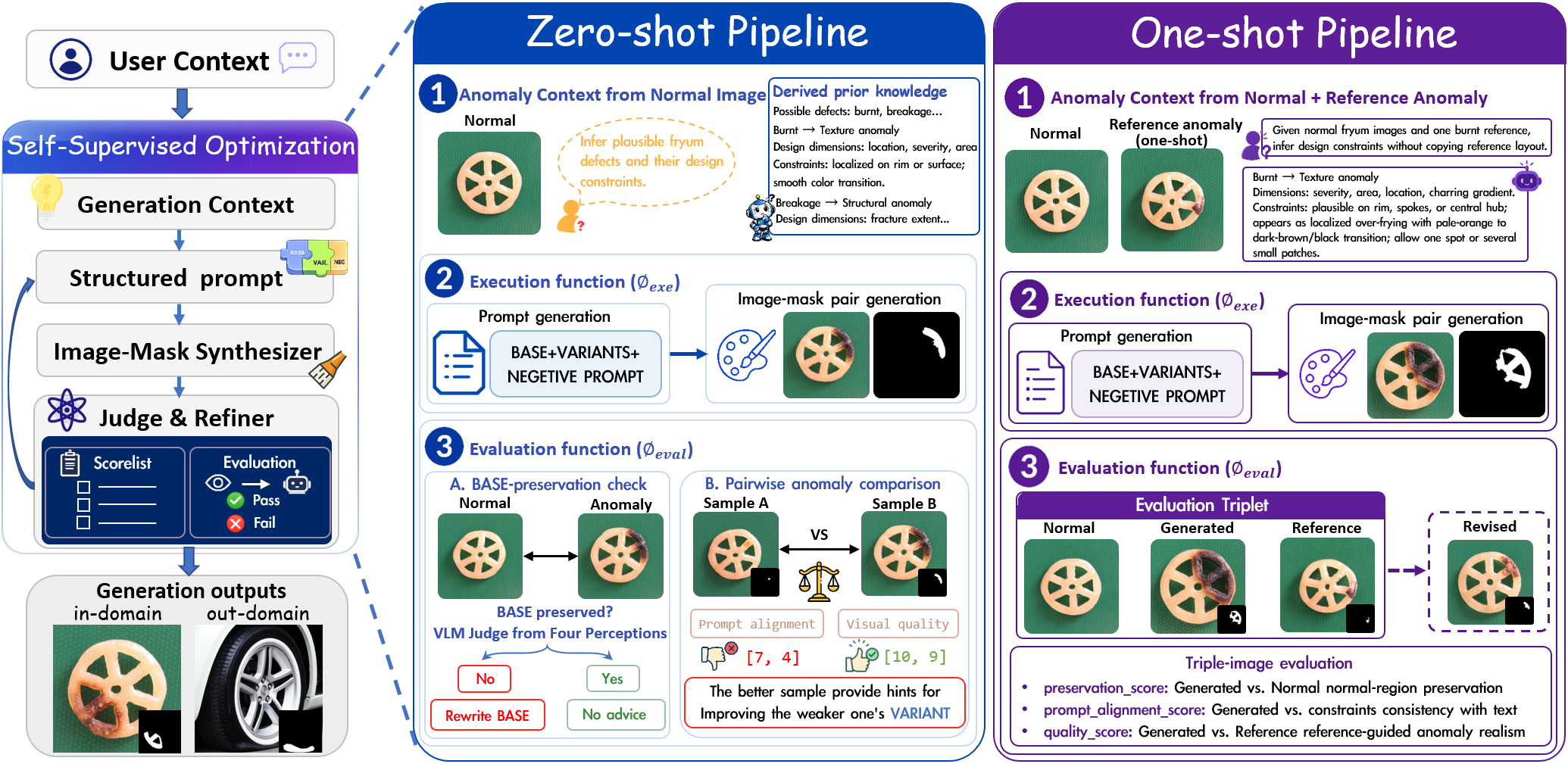}
    \caption{
    \textbf{Overview of the CORE framework.}
    CORE unifies zero-shot and one-shot anomaly synthesis within a self-supervised Execute-Evaluate-Optimize loop. Starting from user-provided inputs, CORE derives anomaly generation context and iteratively evaluates and refines the generation process, enabling realistic and diverse anomaly generation.
    }
    \label{fig:core_framework}
\end{figure*}

\subsection{Framework Overview}
As shown in Fig.~\ref{fig:core_framework}, CORE is a self-supervised prompt optimization framework composed of three fundamental functions: 
(1) Execution function \(\phi_{\mathrm{exe}}\) takes the current prompt \(P\) and invokes the LMM to generate an anomalous image-mask pair \((I^{a},M^{a})\); 
(2) Evaluation function \(\phi_{\mathrm{eval}}\) assesses the quality of the generated anomaly image and produces feedback \(F\), which serves as guidance for subsequent prompt optimization; 
(3) Optimization function \(\phi_{\mathrm{opt}}\) updates the current anomaly prompt according to the feedback \(F\) and feeds the revised prompt \(P^{\star}\) back to \(\phi_{\mathrm{exe}}\) for image-mask regeneration, enabling iterative refinement throughout the optimization process.

These functions form a closed loop of execution, evaluation, and optimization. Starting from an initial prompt, CORE automatically generates anomaly image-mask pairs, diagnoses their failure modes, and refines the structured prompt for the next iteration. Through this feedback-driven refinement process, CORE converts generated results into optimization feedback, enabling iterative prompt refinement with black-box LMMs without manual intervention.

\subsection{Structured Prompt Formulation}
To bridge the context gap between sparse user context and defect-aware generation context, CORE decomposes prompt formulation into two stages. It first derives anomaly design dimensions and generation constraints from the normal images, optionally supplemented by one anomaly reference, and then instantiates them into a structured prompt for image--mask generation and refinement. This formulation reduces the ambiguity of direct prompting and avoids implausible anomalies and unintended changes to normal regions.

\noindent\textbf{Anomaly Context Grounding.}
Given two normal images, optionally supplemented with one anomaly reference in the one-shot setting, CORE derives anomaly generation contexts from two complementary aspects: design dimensions \(D\) and generation constraints \(G\). The design dimensions specify the admissible variation space of the target anomaly, whereas the constraints restrict their instantiation to semantically consistent, structurally plausible, and material-consistent manifestations while preserving normal regions and suppressing irrelevant changes. This separation distinguishes the anomaly variation space from the validity conditions governing generation and defines a constrained anomaly space that provides a unified basis for prompt construction, result evaluation, and prompt optimization. In the zero-shot setting, CORE relies solely on the normal images and leverages the LMM's internal knowledge to infer plausible anomaly variations and object-specific generation constraints. In the one-shot setting, one real anomaly reference further enriches the context with informative visual cues about defect appearance and material interaction, without encouraging instance-level copying.

\begin{figure*}[t]
    \centering
    \includegraphics[
        page=1,
        width=\textwidth
    ]{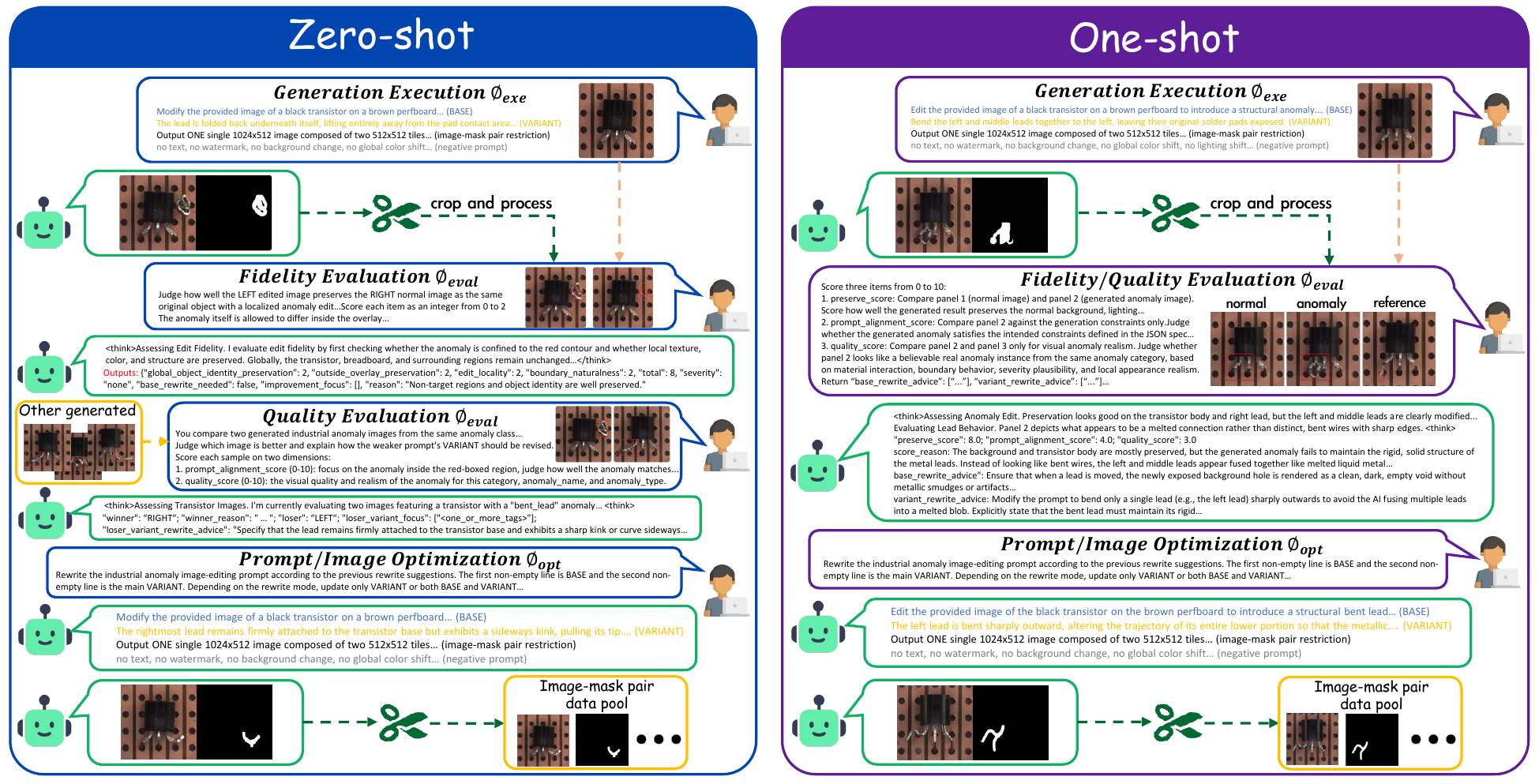}
    \caption{A running example of feedback-driven prompt optimization in CORE. Starting from an initial structured prompt, CORE performs generation, evaluation, and prompt optimization to progressively improve the synthesized image--mask pairs.}
    \label{fig:pipeline}
\end{figure*}

\noindent\textbf{Structured Anomaly Prompt.}
Based on the grounded anomaly context, CORE instantiates the executable anomaly prompt as \(P=\{B,V,N\}\), where \(B\), \(V\), and \(N\) denote \textbf{BASE}, \textbf{VARIANT}, and \textbf{NEGATIVE}, respectively. BASE encodes the preservation constraints in \(G\), specifying the source content that should remain invariant during editing. VARIANT instantiates anomaly-specific attributes from the design dimensions \(D\), including anomaly location, morphology, texture and material interaction, while remaining consistent with the constraints defined in \(G\). NEGATIVE is a general-purpose clause to suppress irrelevant artifacts, global appearance shifts and implausible deformations, and remains fixed throughout optimization. During execution, a fixed mask instruction \(M\) is inserted, enabling joint generation of an aligned anomaly image--mask pair. Through this decomposition, CORE explicitly disentangles normal-region preservation from anomaly semantic variation, reducing editing drift. Moreover, the component-wise prompt structure aligns evaluation feedback with the corresponding prompt component, enabling targeted and interpretable refinement.

\subsection{Self-Supervised Prompt Optimization}
CORE optimizes the structured anomaly prompt by converting generated results into self-supervised feedback signals. The key observation is that prompt quality is implicitly reflected in the generated outputs: a high-quality anomaly prompt should produce defects that are semantically aligned with the intended anomaly, locally realistic for the target material, and faithful to the normal regions of the input image. Instead of manual prompt tuning, CORE identifies generation deficiencies and revises the corresponding prompt components through a contrast-and-reflect mechanism.

Given a normal image \(I^n\) and a structured anomaly prompt \(P=\{B,V,N\}\), the execution function \(\phi_{\mathrm{exe}}\) invokes the black-box LMM to generate an anomaly image-mask pair:
\begin{equation}
    (I^a, M^a) = \phi_{\mathrm{exe}}(I^n, P),
\end{equation}
where \(I^a\) and \(M^a\) denote the synthesized anomalous image and its corresponding anomaly mask. The generated result is then evaluated under a setting-specific feedback function:
\begin{equation}
\small
    F = \phi_{\mathrm{eval}}
    \left(I^n, G, P, (I^a, M^a), \mathcal{R}^{(s)}\right), 
    \quad
    P_{\mathrm{opt}} = \phi_{\mathrm{opt}}(P, F),
\end{equation}
where \(s \in \{\mathrm{zero}, \mathrm{one}\}\) denotes the synthesis setting. Based on the failure modes identified in \(F\), the optimization function selectively rewrites BASE or VARIANT, while the shared \textbf{NEGATIVE} component \(N\) remains unchanged.

\noindent\textbf{Zero-shot.}
In the zero-shot setting, no real anomaly reference is available for assessing anomaly realism. CORE addresses this challenge by deriving self-supervised optimization signals from the relative evaluation of its generated outputs. Given a user-specified number of candidates \(K\), CORE constructs a candidate pool from multiple prompt variants:
\begin{equation}
    \mathcal{C} =
    \left\{
    (P_k, I_k^{a}, M_k^{a})
    \right\}_{k=1}^{K},
\end{equation}
where each candidate is generated from a different structured prompt. Since no ground-truth anomaly is available, CORE decouples the evaluation into normal-region preservation and anomaly-region quality.

For normal-region preservation, CORE compares each generated result with the input normal image:
\begin{equation}
    F_{\mathrm{BASE},k} =
    \phi_{\mathrm{eval}}^{\mathrm{BASE}}
    \left(I_k^{n}, (I_k^{a}, M_k^{a})\right).
\end{equation}
This evaluation focuses on whether regions outside the anomaly mask are faithfully preserved, thereby identifying BASE-related failures such as background changes, lighting shifts, material inconsistency, or object-identity distortion.

For anomaly-region quality, CORE performs pairwise output comparison between generated candidates:
\begin{equation}
    F_{\mathrm{VARIANT}}^{i,j} =
    \phi_{\mathrm{eval}}^{\mathrm{OvO}}
    \left(
    (P_i, I_i^{a}, M_i^{a}),
    (P_j, I_j^{a}, M_j^{a})
    \right), \quad i \neq j.
\end{equation}

Here, \textit{OvO} denotes Output \textnormal{vs.} Output evaluation, which contrasts outputs from different prompts to derive relative quality feedback while accommodating multiple valid defect appearances. The LMM evaluates each pair by \textit{prompt alignment} with \(G\) and \textit{anomaly quality} in terms of local realism, structural plausibility, and material consistency. The preferred candidate provides evidence for identifying deficiencies in the weaker candidate, enabling targeted correction without encouraging appearance-level imitation. Pairwise feedback assigned to each candidate is then aggregated as \(F_{\mathrm{VARIANT},k}\), and the corresponding prompt is refined by
\begin{equation}
    P_{\mathrm{opt},k}
    =
    \phi_{\mathrm{opt}}
    \left(
    P_k,
    F_{\mathrm{BASE},k},
    F_{\mathrm{VARIANT},k}
    \right).
\end{equation}
CORE thus derives reference-free optimization feedback from its generated candidates, enabling zero-shot prompt refinement without real anomaly samples.

\noindent\textbf{One-shot.}
In the one-shot setting, CORE incorporates a single real anomaly reference \((I_{\mathrm{ref}},M_{\mathrm{ref}})\) as weak visual guidance. Rather than serving as a template for direct replication, the reference provides evidence about defect appearance, structural plausibility, and material interaction. Given a generated result \((I^a,M^a)\), CORE performs Output vs. Ground Truth (\textit{OvG}) evaluation by jointly assessing the source normal image, generated anomaly and real anomaly reference:
\begin{equation}
    F_{\mathrm{OvG}} =
    \phi_{\mathrm{eval}}^{\mathrm{OvG}}
    \left(
    I^n,
    G,
    P,
    (I^a,M^a),
    (I_{\mathrm{ref}},M_{\mathrm{ref}})
    \right).
\end{equation}
The LMM evaluates the generated result along three dimensions: \textit{preservation score}, measuring the fidelity of normal regions; \textit{prompt-alignment score}, measuring consistency with the textual constraints $G$; and \textit{quality score}, assessing anomaly realism under the guidance of the real reference. Together, these scores and their diagnostic feedback constitute \(F_{\mathrm{OvG}}\), which determines whether the current prompt is retained or whether \textsc{Base}, \textsc{Variant}, or both are refined. The optimized prompt is obtained as
\begin{equation}
    P_{\mathrm{opt}} =
    \phi_{\mathrm{opt}}(P,F_{\mathrm{OvG}}).
\end{equation}
Consequently, CORE derives feedback directly from generated outputs in the zero-shot setting and incorporates weak reference guidance in the one-shot setting, providing a unified formulation for anomaly synthesis. Detailed scoring and refinement rules are provided in the appendix.

\section{Experiments}

\begin{table*}[t]
\centering
\small
\setlength{\tabcolsep}{4pt}
\renewcommand{\arraystretch}{1.08}
\resizebox{\textwidth}{!}{
\begin{tabular}{l|ccccccc|ccccccc}
\toprule
\multicolumn{1}{c|}{} 
& \multicolumn{7}{c|}{\textbf{MVTec-AD}} 
& \multicolumn{7}{c}{\textbf{VisA}} \\
\cmidrule(lr){2-8} \cmidrule(lr){9-15}
\textbf{Method} 
& \textbf{I-AUC} & \textbf{I-AP} & \textbf{I-F1} 
& \textbf{P-AUC} & \textbf{P-AP} & \textbf{P-F1} & \textbf{PRO}
& \textbf{I-AUC} & \textbf{I-AP} & \textbf{I-F1} 
& \textbf{P-AUC} & \textbf{P-AP} & \textbf{P-F1} & \textbf{PRO} \\
\midrule
\multicolumn{15}{c}{\textit{\textbf{One-Shot Anomaly Generation}}} \\
\midrule
\textbf{TF-IDG~\cite{xu2025training}}
& 89.9 & 95.6 & 93.0 & 88.5 & 48.6 & 50.2 & 76.2
& 90.9 & 93.1 & 86.9 & 93.7 & 44.3 & 47.4 & 80.6 \\

\textbf{O2MAG~\cite{rao2026one}}
& \underline{95.9} & \underline{98.3} & \underline{94.9} 
& \underline{89.4} & \underline{51.2} & \underline{51.7} & \underline{83.2}
& \underline{91.7} & \underline{94.1} & \underline{88.1} 
& \underline{94.8} & \underline{47.5} & \underline{50.7} & \underline{83.8} \\

\textbf{CORE (Ours)}
& \textbf{96.5} & \textbf{98.7} & \textbf{96.3} 
& \textbf{96.5} & \textbf{66.7} & \textbf{64.6} & \textbf{87.2}
& \textbf{92.5} & \textbf{94.5} & \textbf{88.7} 
& \textbf{97.9} & \textbf{53.0} & \textbf{55.1} & \textbf{84.9} \\
\midrule
\multicolumn{15}{c}{\textit{\textbf{Zero-Shot Anomaly Generation}}} \\
\midrule
\textbf{DRAEM (Zavrtanik et al.~\citeyear{zavrtanik2021draem})}
& \underline{92.2} & 96.2 & \underline{92.5} & 85.5 & 43.4 & 45.1 & 74.2
& 85.7 & 87.8 & 82.9 & 89.7 & 20.0 & 26.4 & 75.9 \\

\textbf{AnomalyAny~\cite{sun2025unseen}}
& 86.8 & 94.1 & 89.8 & \underline{89.7} & 37.1 & 40.8 & 74.7
& 83.5 & 85.6 & 82.5 & \underline{90.3} & 17.7 & 22.8 & 72.8 \\

\textbf{Anomagic~\cite{jiang2026anomagic}}
& 92.0 & \underline{96.4} & 92.4 
& 88.1 & \underline{44.5} & \underline{46.9} & \underline{79.5}
& 87.5 & 90.0 & 83.8 & 89.2 & 20.7 & 27.5 & 75.9 \\

\textbf{AnoStyler~\cite{so2026anostyler}}
& 89.1 & 95.1 & 91.5 
& 82.7 & 37.5 & 39.2 & 71.1
& \underline{88.4} & \underline{92.0} & \underline{85.3} 
& 89.9 & \underline{23.8} & \underline{30.4} & \underline{82.5} \\

\textbf{CORE (Ours)}
& \textbf{96.4} & \textbf{98.9} & \textbf{95.7} 
& \textbf{96.4} & \textbf{62.8} & \textbf{61.6} & \textbf{87.4}
& \textbf{90.9} & \textbf{92.3} & \textbf{86.7} & \textbf{96.2} & \textbf{38.5} & \textbf{44.3} & \textbf{82.6} \\
\bottomrule
\end{tabular}
}
\caption{Comparison of one-shot and zero-shot anomaly synthesis methods on MVTec-AD and VisA. All results are reported as percentages, with the best and second-best results within each setting highlighted in \textbf{bold} and \underline{underlined}, respectively.}
\label{tab:main_results}
\end{table*}

\subsection{Experiment Settings}
\noindent\textbf{Datasets.}
We conduct experiments on two representative industrial anomaly detection benchmarks: MVTec-AD~\cite{bergmann2019mvtec} and VisA~\cite{zou2022spot}. MVTec-AD contains 10 object categories and 5 texture categories, while VisA includes 12 categories of complex industrial objects. To simulate realistic industrial inspection scenarios with scarce anomalies, we evaluate CORE under two settings: (1) \textbf{zero-shot setting}, where no anomalous samples are used for synthesis; and (2) \textbf{one-shot setting}, where one anomalous sample is provided as the reference for each defect type and excluded from downstream evaluation, ensuring no overlap between the reference and evaluation sets.

\begin{table}[t]
\centering
\small
\setlength{\tabcolsep}{3pt} 
\renewcommand{\arraystretch}{0.85} 
\resizebox{\linewidth}{!}{
\begin{tabular}{l|cc|cc}
\toprule
\multicolumn{1}{c|}{}
& \multicolumn{2}{c}{MVTec-AD}
& \multicolumn{2}{c}{VisA} \\
\cmidrule(lr){2-3} \cmidrule(lr){4-5}
Method
& IC-L $\uparrow$ & IS $\uparrow$
& IC-L $\uparrow$ & IS $\uparrow$ \\
\midrule
\multicolumn{5}{c}{Zero-Shot Anomaly Generation} \\
\midrule
AnomalyAny~\cite{sun2025unseen}  & \textbf{0.29} & 2.04 & 0.10 & \underline{1.66} \\
Anomagic~\cite{jiang2026anomagic}    & \underline{0.28} & \underline{2.09} & \textbf{0.34} & \textbf{1.83} \\
AnoStyler~\cite{so2026anostyler}   & 0.27 & 1.59 & 0.31 & 1.37 \\
CORE (Ours) & \underline{0.28} & \textbf{2.13} & \textbf{0.34} & 1.54 \\
\midrule
\multicolumn{5}{c}{One-Shot Anomaly Generation} \\
\midrule
TF-IDG~\cite{xu2025training}      & 0.23 & \underline{1.73} & 0.25 & 1.32 \\
O2MAG~\cite{rao2026one}       & \underline{0.27} & 1.56 & \underline{0.31} & \underline{1.41} \\
CORE (Ours) & \textbf{0.28} & \textbf{2.08} & \textbf{0.33} & \textbf{1.51} \\
\bottomrule
\end{tabular}
}
\caption{Comparison of anomaly generation on MVTec-AD and VisA under zero-shot and one-shot anomaly synthesis settings. \textbf{Bold} and \underline{underline} represent optimal and sub-optimal results. On VisA, some zero-shot methods introduce larger changes at implausible locations, which may artificially inflate their IS.}
\label{tab:icl_is_results}
\end{table}

\noindent\textbf{Baselines.}
We evaluate the training-free CORE framework under both zero-shot and one-shot settings. In the zero-shot setting, we compare CORE with DRAEM~\cite{zavrtanik2021draem} (handcrafted synthesis), AnomalyAny~\cite{sun2025unseen} (training-free method), Anomagic~\cite{jiang2026anomagic} (fine-tuning with inpainting Stable Diffusion model), and AnoStyler~\cite{so2026anostyler} (fine-tuning with stylization network). In the one-shot setting, we compare CORE with recent training-free methods TF-IDG~\cite{xu2025training} and O2MAG~\cite{rao2026one}. Training-based few-shot methods are excluded because they are designed for few-shot protocols, typically using one-third of the anomalous images in each defect category as reference data for generation. Under the stricter one-shot setting, their training process becomes unstable and prone to overfitting.

\noindent\textbf{Implementation Details.}
We use Gemini-3.1-Pro as the text-side agent for prompt generation, evaluation, and optimization, and Nano Banana 2 as the image generator. To further validate the generality of CORE, we also adopt OpenAI's GPT-5 Mini and GPT-Image-2 for prompt optimization and image synthesis in the ablation study. For cost efficiency, we perform one iteration of the Execute-Evaluate-Optimize loop, which already yields substantial gains in downstream anomaly detection. Fig.~\ref{fig:pipeline} illustrates an example of CORE's optimization process. The text-side stages average 9 seconds and \$0.01 per sample, while image generation requires $18$ seconds/\$0.06 with Nano Banana 2 and $15$ seconds/\$0.03 with GPT-Image-2. More details are supplied in Appendix~B.

\begin{figure*}[t]
    \centering
    \includegraphics[width=\textwidth]{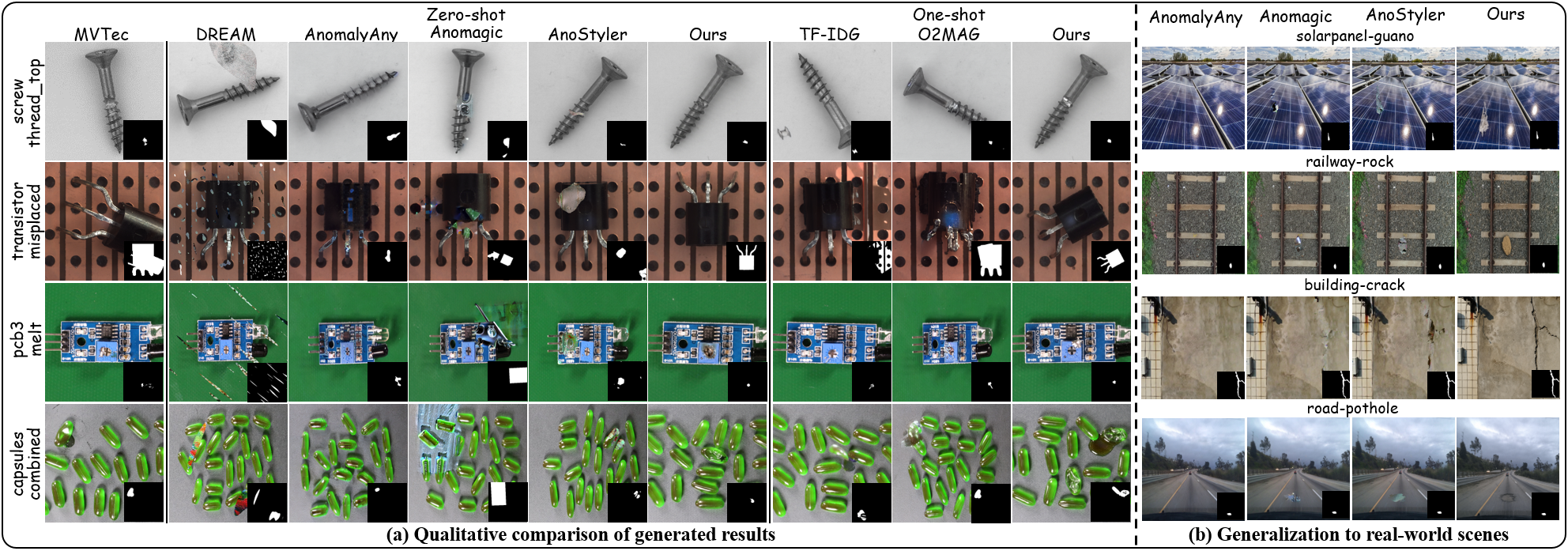}
    \caption{
    \textbf{Qualitative comparison of anomaly synthesis results.}
    (a) compares CORE with representative zero-shot and one-shot anomaly synthesis methods on MVTec-AD and VisA, showing that CORE produces more realistic and localized anomalies while better preserving non-anomalous regions; (b) further illustrates its generalization to real-world scenes.
    }
    \label{fig:qualitative_comparison}
\end{figure*}

\begin{table*}[t]
\centering
\small
\setlength{\tabcolsep}{3.4pt}
\renewcommand{\arraystretch}{0.82}
\resizebox{\textwidth}{!}{
\begin{tabular}{ll|ccccccc|ccccccc}
\toprule
\multicolumn{1}{c}{\textbf{Setting}}
& \multicolumn{1}{c|}{\textbf{Variant}}
& \multicolumn{7}{c|}{\textbf{MVTec-AD}}
& \multicolumn{7}{c}{\textbf{VisA}} \\
\cmidrule(lr){3-9} \cmidrule(lr){10-16}
&
& \textbf{I-AUC} & \textbf{I-AP} & \textbf{I-F1}
& \textbf{P-AUC} & \textbf{P-AP} & \textbf{P-F1} & \textbf{PRO}
& \textbf{I-AUC} & \textbf{I-AP} & \textbf{I-F1}
& \textbf{P-AUC} & \textbf{P-AP} & \textbf{P-F1} & \textbf{PRO} \\
\midrule

\multirow{4}{*}{One-shot}
& w/o ACG
& 93.2 & 97.2 & 93.8 & 94.7 & 59.7 & 58.6 & 85.1
& 88.6 & 91.0 & 85.3 & 96.2 & 43.1 & 45.7 & 78.4 \\
& Initial (ACG)
& 94.2 & 97.9 & 94.6 & 95.3 & 63.5 & 61.6 & 85.2
& 91.3 & 93.2 & 87.3 & 96.0 & 45.1 & 49.3 & 83.0 \\
& Google
& 96.5 & 98.7 & 96.3 & 96.5 & 66.7 & 64.6 & 87.2
& 92.5 & 94.5 & 88.7 & 97.9 & 53.0 & 55.1 & 84.9 \\
& OpenAI
& 95.3 & 98.3 & 95.3 & 97.2 & 67.4 & 64.4 & 88.3
& 91.8 & 94.1 & 88.3 & 97.5 & 50.0 & 53.6 & 86.5 \\

\midrule

\multirow{4}{*}{Zero-shot}
& w/o ACG
& 93.1 & 97.3 & 94.2 & 95.2 & 57.6 & 57.3 & 83.7
& 88.6 & 90.6 & 84.9 & 95.3 & 35.0 & 40.2 & 80.5 \\
& Initial (ACG)
& 93.0 & 97.7 & 93.9 & 95.4 & 60.1 & 59.0 & 84.5
& 88.8 & 91.3 & 85.0 & 95.5 & 36.8 & 42.4 & 81.4 \\
& Google
& 96.4 & 98.9 & 95.7 & 96.4 & 62.8 & 61.6 & 87.4
& 90.9 & 92.3 & 86.7 & 96.2 & 38.5 & 44.3 & 82.6 \\
& OpenAI
& 95.2 & 98.5 & 95.7 & 96.1 & 66.0 & 63.2 & 87.4
& 89.3 & 92.0 & 86.0 & 96.0 & 41.6 & 44.5 & 83.8 \\

\bottomrule
\end{tabular}
}
\caption{Ablation study of ACG and prompt optimization with different image generators. \textit{Initial (ACG)} is shared by the Google and OpenAI optimization
backends, while \textit{w/o ACG} removes anomaly context grounding.}
\label{tab:ablation_prompt_lmm}
\end{table*}

\subsection{Anomaly Generation Results}
Following the full-shot experimental setting of AnomalyAny, we generate 100 anomalous images for each defect type. DRAEM~\cite{zavrtanik2021draem} is excluded because it synthesizes anomalies using random textures cropped from external datasets. We adopt Inception Score (IS)~\cite{salimans2016improved} and Intra-Cluster pairwise LPIPS distance (IC-L)~\cite{ojha2021few} to evaluate generation quality and diversity, respectively.

As shown in Table~\ref{tab:icl_is_results}, CORE achieves the best performance across all metrics in the one-shot setting, demonstrating that it effectively leverages a single anomaly reference as a weak visual prior while preserving generation diversity and avoiding appearance-level overfitting. CORE also achieves competitive performance in the zero-shot setting. Anomagic obtains the highest class-wise IS on VisA, whereas the examples in Fig.~\ref{fig:qualitative_comparison}(a) reveal that AnomalyAny introduces unintended changes to the board texture and component appearance, and Anomagic generates a prominent metallic structure extending beyond the local component region. Since IS evaluates full-image statistics, such salient changes may improve the score without necessarily enhancing localized defect fidelity. Additionally, existing zero-shot methods often struggle to synthesize complex logical anomalies, whereas one-shot baselines remain sensitive to the spatial plausibility of anomaly masks. By contrast, CORE better preserves normal content while generating content-aware logical anomalies and structurally consistent image-mask pairs. Fig.~\ref{fig:qualitative_comparison}(b) further demonstrates its generalization to unseen objects in open-world scenarios, with additional qualitative examples provided in Appendix~E.

\subsection{Anomaly Detection Results}
We further evaluate the downstream effectiveness of different anomaly synthesis methods for anomaly detection and localization, following the evaluation protocols of O2MAG~\cite{rao2026one} and TF-IDG~\cite{xu2025training}. For each defect type, we synthesize 100 anomalous images and use them to train a U-Net segmentation model that predicts pixel-level anomaly masks from input images. We report Area Under ROC Curve (AUROC), Average Precision (AP), and F1-score at both the image level and pixel level. We additionally report Per-Region Overlap (PRO) to assess region-level anomaly localization. Table~\ref{tab:main_results} reports the results averaged over categories within each benchmark dataset. The full category-wise results of CORE are provided in Appendix~D. Under both zero-shot and one-shot settings, CORE achieves state-of-the-art anomaly detection performance on MVTec-AD and VisA, highlighting its strong generalization ability.

\subsection{Ablation Study}
Table~\ref{tab:ablation_prompt_lmm} evaluates the effectiveness of ACG and prompt optimization, along with CORE's cross-backend generalization.

\noindent\textbf{Effect of Anomaly Context Grounding.}
We compare ACG with direct prompting under the Google configuration with Gemini-3.1-Pro and Nano Banana 2, while keeping the visual inputs and all other experimental settings unchanged. As shown in Table~\ref{tab:ablation_prompt_lmm}, ACG consistently improves performance by transforming sparse user input into the defect-aware generation context required by general-purpose LMMs.

\noindent\textbf{Effect of Prompt Optimization.}
Starting from this shared ACG initialization, the Google configuration consistently improves downstream anomaly detection and localization through prompt optimization. These gains validate the effectiveness of the proposed contrast-and-reflect mechanism in refining anomaly prompts and enhancing the downstream utility of synthesized samples.

\noindent\textbf{Generalization across Model Backends.}
We further replace the optimization-stage LMM and regeneration backend with GPT-5 Mini and GPT-Image-2 while keeping the shared initialization fixed. The resulting improvements under both zero-shot and one-shot settings demonstrate its model-agnostic generalization across model backends. Detailed category-wise results are provided in Appendix~D.

\section{Conclusion}
In this paper, we introduced \textbf{CORE}, a cost-efficient and training-free prompt optimization framework that unifies zero-shot and one-shot industrial anomaly synthesis. Through Anomaly Context Grounding, CORE transforms sparse user-provided context into the defect-aware generation context required by general-purpose LMMs and produces structured prompts that disentangle normal-content preservation from anomaly-specific variation.
CORE further enables automatic prompt optimization through a self-supervised contrast-and-reflect mechanism, without task-specific training or manual prompt tuning.
Extensive experiments on MVTec-AD and VisA demonstrate that CORE improves both generation quality and downstream anomaly detection performance. Its model-agnostic design further enables plug-and-play transfer across different LMM backends and generalization to unseen objects in open-world scenarios.

\bibliography{aaai2027}




\end{document}